# JOINT SUPER-RESOLUTION AND RECTIFICATION FOR SOLAR CELL INSPECTION

*Mathis Hoffmann*[⋆†]   *Thomas Köhler*[∥]   *Bernd Doll*[†§¶]
*Frank Schebesch*[⋆]   *Florian Talkenberg*[‡]   *Ian Marius Peters*[§]
*Christoph J. Brabec*[†¶]   *Andreas Maier*[⋆¶]   *Vincent Christlein*[⋆]

[⋆] Pattern Recognition Lab, University of Erlangen-Nürnberg (FAU), Germany
[†] Institute Materials for Electronics and Energy Technology, FAU, Germany
[‡] greateyes GmbH, Berlin, Germany
[§] Forschungszentrum Jülich GmbH, Helmholtz Institut Erlangen-Nürnberg, Germany
[¶] Graduate School in Advanced Optical Technologies, Erlangen, Germany
[∥] e.solutions GmbH, Erlangen, Germany

## ABSTRACT

Visual inspection of solar modules is an important monitoring facility in photovoltaic power plants. Since a single measurement of fast CMOS sensors is limited in spatial resolution and often not sufficient to reliably detect small defects, we apply multi-frame super-resolution (MFSR) to a sequence of low resolution measurements. In addition, the rectification and removal of lens distortion simplifies subsequent analysis. Therefore, we propose to fuse this pre-processing with standard MFSR algorithms. This is advantageous, because we omit a separate processing step, the motion estimation becomes more stable and the spacing of high-resolution (HR) pixels on the rectified module image becomes uniform w. r. t. the module plane, regardless of perspective distortion. We present a comprehensive user study showing that MFSR is beneficial for defect recognition by human experts and that the proposed method performs better than the state of the art. Furthermore, we apply automated crack segmentation and show that the proposed method performs $3\times$ better than bicubic upsampling and $2\times$ better than the state of the art for automated inspection.

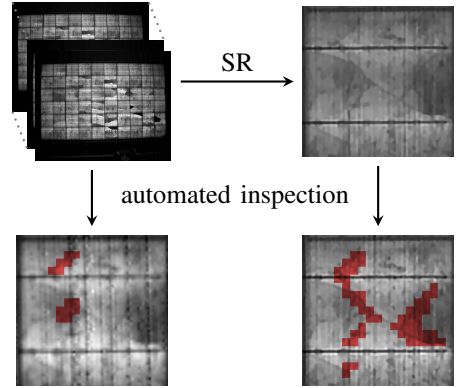

**Fig. 1**: We apply super-resolution (SR) to a sequence of low resolution images of photovoltaic modules and show that the resulting images (right) are better suited for the inspection by human experts as well as automated inspection, compared to the original low-resolution (LR) images (left).

## I. INTRODUCTION

VISUAL inspection using imaging technologies is an essential part of material quality control. For example, solar modules are subject to regular inspection by electroluminescence imaging [1] for material defect detection. To inspect every single cell of a solar module, the entire module needs to be located in the electroluminescence (EL) image first, to allow for a subsequent segmentation into single cells (see Fig. 1). In addition, removal of perspective and lens distortion simplifies further analysis [2]. Modern CMOS sensors allow EL imaging at high frame rates, which is beneficial for drone-based in-field inspection. However, due to practical requirements on the signal-to-noise ratio and the sensor integration time, their spatial resolution for EL imaging is limited, challenging the detection of small material defects. In this respect, the use of super-resolution (SR) techniques for software-based resolution enhancement is promising.

In this work, we propose to capture a sequence of LR images using fast CMOS sensors and to subsequently apply multi-frame super-resolution (MFSR) in order cell images at higher resolution. Prior to the reconstruction of a HR image, the sequence of LR images needs to be registered. Traditionally, one of the frames is used as a reference frame that all others are registered to. However, this approach does not compensate for perspective or lens distortion. We propose to integrate the rectification and undistortion steps into the motion estimation procedure of a MFSR approach by performing registration w. r. t. a virtual reference of the module.

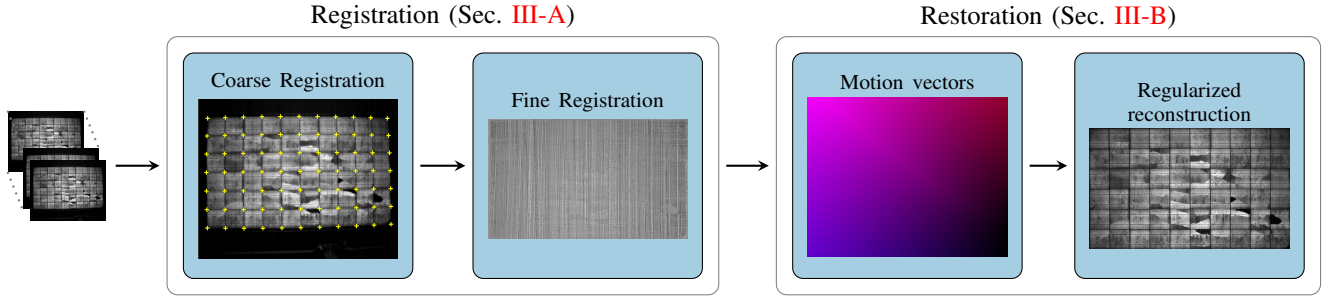

Fig. 2: Overview of the registration and super-resolution framework. First, *coarse registration* of modules is performed by detecting the cell crossing points in every low resolution frame. Then, modules are registered in a virtual coordinate frame that is free of perspective or lens distortion during *fine registration*. Next, *motion vectors* are computed. This results in a motion vector field that relates every LR pixel with the corresponding high-resolution (HR) pixel in the virtual coordinate frame. Here, we color-code the direction, whereas the magnitude corresponds to the brightness. Finally, the distortion-free HR image is obtained by *regularized reconstruction*.

The contributions of our work are as follows:
1) We fuse rectification and undistortion with SR which leads to a uniform spacing of pixels in the module plane, regardless of perspective or lens distortion.
2) We conduct an extensive user study showing that MFSR simplifies manual inspection of solar cells by human experts and that our proposed registration outperforms the state of the art.
3) We apply automated crack detection and segmentation on the resulting images and show that MFSR enhances automated inspection of solar cells.

## II. RELATED WORK

Visual inspection of solar cells is an active research area. In previous studies, it has been shown that, for example, micro cracks result in power loss of the module, especially after simulated aging [3]. Hence, reliable detection of these and other defect types is of particular interest. Only recently, deep learning (DL)-based methods are applied to automatically detect defects on solar cells using EL imaging [4]. Many of these works focus on the detection [5] and segmentation [6] of cracks. However, a reliable detection of small cracks is challenging and requires a sufficient image resolution. For this reason, we focus on the automated detection and segmentation of cracks and show that super-resolution is beneficial for this task.

**Observation model for super-resolution.** Given a sequence of LR images captured from a scene of interest, SR can be modelled by the observation model:

$$\boldsymbol{g}_i = \boldsymbol{W}_i \boldsymbol{f} + \boldsymbol{n}_i \quad \text{for} \quad 1 \leq i \leq N \, , \qquad (1)$$

where $\boldsymbol{f}$ denotes the desired HR image, $\boldsymbol{g}_i$ is the $i$-th frame of the LR sequence, and $\boldsymbol{n}_i$ is additive noise. The system matrix $\boldsymbol{W}_i$ describes the imaging process including motion between the HR and LR domain, the point spread function of the optical system, and sampling on the sensor array. SR algorithms aim at inverting this model w. r. t. $\boldsymbol{f}$.

**Single-frame super-resolution.** Many mainstream methods use single images ($N = 1$) to compute a super-resolved image . Here, especially deep neural networks [7], [8], [9], [10] and generative adversarial learning [11], [12] advanced the state of the art in single-frame super-resolution (SFSR). They greatly benefit from large natural image databases, enabling supervised learning of mappings from LR to the HR domain. However, without fine-tuning to the target application, their performance can drastically drop. This is particularly severe in imaging applications such as ours, where comprehensive training datasets are still unavailable and image properties like noise or blur models can considerably deviate from those of natural images [13]. To overcome this issue, zero-short SFSR methods have been developed [14], [15]. These methods exploit recurring information within a single image to train a convolutional neural network at test time. Only recently, it has been shown that combining classification and SFSR into a single network in terms of multi-task learning helps to improve the performance of SFSR and classification at the same time [16]. Still, existing SFSR methods mostly target to satisfy quality perceptions of humans and might even hallucinate HR additional details. This is caused by the fact that they only use the information available in a single image and incorporate prior knowledge to reconstruct HR the image. In contrast, visual inspection requires authentic data for quantitative and objective analyses, like solar cell defect detection. This prohibits the simple use of existing pre-trained models.

**Image-based multi-frame super-resolution.** MFSR aims at the inversion of Eq. (1) using multiple frames ($N > 1$) assuming sub-pixel inter-frame motion, to fuse them into an HR image. This can be done by non-uniform interpolation [17], [18], iterative reconstruction to invert Eq. (1) [19], [20], [21], [22], [23], [24], or deep learning [25], [26], [27], [28], [29], [30], [31] to directly predict the HR image.

During the imaging process, crucial details are lost in LR frames, which is mainly caused by the sampling process on the sensor. In principle, MFSR is able to reconstruct crucial HR details on solar cells from aliasing artifacts in the LR images [32]. However, such approaches are entirely image-based. Specifically, inter-frame motion is directly obtained on the image plane by classical image-to-image registration or learned motion estimation modules. We show that the robustness of these image-based methods for electroluminescence imaging of solar cells is strongly limited by the accuracy of motion estimation. In contrast, we propose to incorporate 3D geometry into the underlying model to increase robustness of MFSR.

**3D geometry-based multi-frame super-resolution.** In a closely related work to ours, Park and Lee [33] have coupled the estimation of the 3D camera pose, depth, and HR data from LR frames via joint optimization based on a projective model. Similar to this approach, we consider 3D camera motion in a static scene. Additionally, we exploit the fact that imaged solar cells are planar and distortion-free objects in space. Therefore, a purely projective model is not sufficient and we propose to include a pinhole camera model along with radial distortion, such that the overall model complies to the imaging process. This allows a joint rectification and SR in electroluminescence imaging of solar cells. We also show how to seamlessly integrate our general-purpose model in regularized reconstruction based MFSR.

## III. METHODOLOGY

An overview of our method is shown in Fig. 2. It consists of two steps: First, LR images are registered (Sec. III-A). As a result, we know the motion $m_i$ from the reference to every low resolution frame $g_i$. As opposed to previous works on SR, this reference is not an arbitrary low resolution frame. We perform registration w.r.t. to an image of the module that is free of lens and perspective distortion and find that this increases registration stability. Hence, $m_i$ describes the deformation of $f$ by perspective projection and lens distortion (see Fig. 3). To simplify notation, we define $m$ such that it includes the downsampling as well. Hence, $m$ is not a pure motion model. In a second step, this virtual module image is reconstructed from the LR images using the estimated motion (Sec. III-B).

### III-A. Registration

In the general case, $m$ may be any motion and downsampling model that relates two corresponding points $\boldsymbol{x}_{i,k}$ and

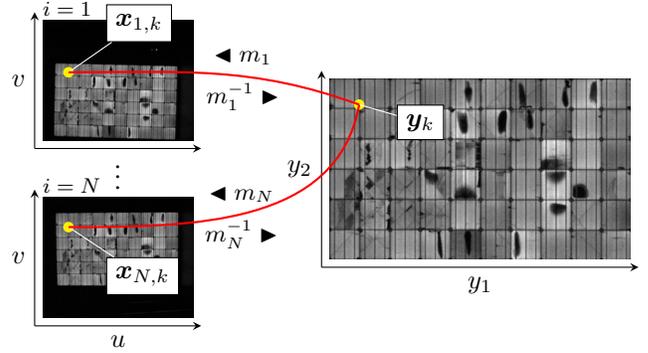

**Fig. 3**: The proposed motion and downsampling model $m_i$ (Eq. (2)) relates every high resolution pixel $\boldsymbol{y}_k = [y_{1,k}, y_{2,k}]$ with the corresponding pixel in the $i$'th low resolution image $\boldsymbol{x}_{i,k} = [u_{i,k}, v_{i,k}]$. Using the inverse $m_i^{-1}$, the HR pixel location corresponding to $\boldsymbol{x}_{i,k}$ is computed, which paves the way for inverse warping.

$\boldsymbol{y}_k$ in either view. Hence, it holds that

$$\boldsymbol{x}_{i,k} = m_i(\boldsymbol{y}_k),  \qquad (2)$$

where $\boldsymbol{x}_{i,k}$ is the LR correspondence to $\boldsymbol{y}_k$ and $k$ denotes the HR pixel index. We assume that the virtual reference of solar modules lies in the $x$-$y$-plane of the reference coordinate system, which we refer to as the module coordinate system. A 2D point in the module coordinate system in homogeneous coordinates $\tilde{\boldsymbol{y}} = [\tilde{y}_1, \tilde{y}_2, s]$ with scaling factor $s$ and its image $\tilde{\boldsymbol{x}}_i = [\tilde{u}_i, \tilde{v}_i, s_i]$ in multiple views $1 \leq i \leq N$ are related by the non-linear model [34], [35]:

$$m_i := \tilde{\boldsymbol{x}}_i \simeq \boldsymbol{K} d(\boldsymbol{H}_i \tilde{\boldsymbol{y}}, \boldsymbol{p}).  \qquad (3)$$

Here, $\boldsymbol{H} \in \mathbb{R}^{3\times 3}$ encodes the camera pose, $\boldsymbol{K}$ are the intrinsic camera parameters, and $d(\cdot, \boldsymbol{p})$ is a distortion model with parameters $\boldsymbol{p}$. We assume that $\boldsymbol{H}$ changes between frames, whereas $\boldsymbol{K}$ and $\boldsymbol{p}$ are constant. For simplicity, we resort to radial distortion with a single coefficient. We further assume that the center of distortion is located at the origin. Let $\tilde{\boldsymbol{y}}' = \boldsymbol{H}\tilde{\boldsymbol{y}} = [\tilde{y}'_1, \tilde{y}'_2, s']$, $y'_1 = \frac{\tilde{y}'_1}{s'}$ and $y'_2 = \frac{\tilde{y}'_2}{s'}$. Then, $d$ follows as

$$d(\tilde{\boldsymbol{y}}', \boldsymbol{p}) = \left(\frac{r_d}{r_u} y'_1, \frac{r_d}{r_u} y'_2\right)^\top,  \qquad (4)$$

where $r_u$ is the distance to the center of distortion, $\kappa$ is the distortion coefficient and $r_d$ is given by [36]

$$r_d = r_u + \kappa r_u^3.  \qquad (5)$$

For registration, we use the approach by Hoffmann *et al.* to find corresponding points between LR frames [37] and Zhang's method to initially estimate the intrinsic and extrinsic camera parameters [38]. Then, registered images are obtained by warping each LR frame into the common module coordinate system. To achieve sub-pixel registrations in the presence of noisy point correspondences, we refine the initial parameters by non-linear least-squares minimization:

$$\hat{\boldsymbol{H}}_i = \arg\min_{\boldsymbol{H}_i} \|\boldsymbol{g}_i - \tau(\boldsymbol{f}, m_i)\|_2. \quad (6)$$

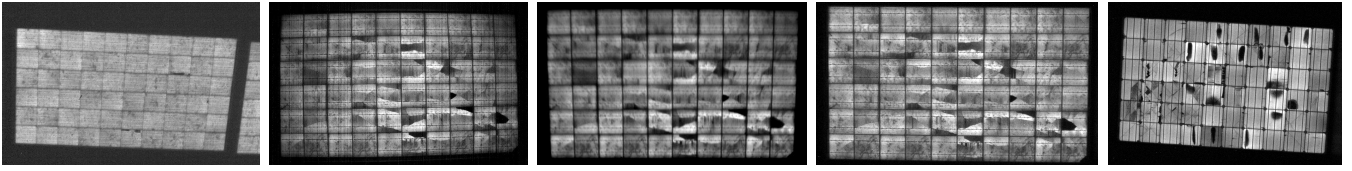

**Fig. 4**: Example images from the datasets D1-5.

Here $\tau(\boldsymbol{f}, m_i)$ denotes inverse warping of $\boldsymbol{f}$ according to the motion model $m_i$. To improve stability, we perform registration in a multi-scale scheme [39].

### III-B. Joint super-resolution and rectification

By the observation model in Eq. (1), the HR image is transformed to obtain a LR view and we can relate any point in the module coordinate frame $\boldsymbol{y}_k$ with its image $\boldsymbol{x}_{i,k}$ in the $i$-th LR frame by Equation (2) and (3) (cf. Fig. 3). However, for smooth results, it is preferable to implement Eq. (1) by inverse warping. Therefore, we need to relate $\boldsymbol{x}_{i,l}$ with its image in $\boldsymbol{y}$, where $l$ is the LR pixel index. Hence, the inverse of $m$ is required. Using the motion model in Eq. (3), we need to invert $\boldsymbol{K}$, $\boldsymbol{H}_i$ and $\boldsymbol{p}$. Since $\boldsymbol{K}$ and $\boldsymbol{H}_i$ are square matrices and have full rank, they are invertible right away. According to Eq. (5), distorted and undistorted radii are related by a polynomial of degree 3, which has an analytic inverse by the formula of Cardano. The solution depends on the discriminant $D$. For $D > 0$, the undistorted radius $r_u$ is given by

$$r_u = \frac{1}{6\kappa} \left( \sqrt[3]{-4q + 4\sqrt{q^2 + 4p^3}} + \sqrt[3]{-4q - 4\sqrt{q^2 + 4p^3}} \right),$$

for $D < 0$ it is

$$r_u = 6\kappa\sqrt{-p} \cos\left(\frac{1}{3}\arccos\left(\frac{-q}{2\sqrt{-p^3}}\right) + \frac{4\pi}{3}\right)$$

and for $D = 0$ it holds that $r_u = r_d$. Note that we substituted $p = 3\kappa$ and $q = -27\kappa^2 r_d$ and that the discriminant is $D = q^2 + 4p^3$. A detailed derivation of this result is given in Sec. A.

Inter-frame motion can be expressed in various forms and we use motion vector fields as a general model. The motion vector for the $l$'th pixel of the $i$'th LR frame is computed as $\boldsymbol{v}_{i,l} = m^{-1}(\boldsymbol{x}_{i,l}) - \boldsymbol{x}_{i,l}$. Using the derived motion fields, a known camera point spread function, and the desired SR magnification factor, we parameterize the system matrices $\boldsymbol{W}_i, i = 1, \ldots, N$ of the observation model in Eq. (1) similar to [21]. Then, $\boldsymbol{W}_i$ relates the HR, undistorted and rectified image $\boldsymbol{f}$ of the virtual module to the LR frame $\boldsymbol{g}_i$. Given $N$ frames, we aim at reconstructing $\boldsymbol{f}$ as maximum aposteriori estimate:

$$\hat{\boldsymbol{f}} = \arg\min_{\boldsymbol{f}} \sum_{i=1}^{N} \|\boldsymbol{g}_i - \boldsymbol{W}_i \boldsymbol{f}\|_2^2 + \lambda R(\boldsymbol{f}), \quad (7)$$

where $R(\boldsymbol{f})$ denotes a regularization term with regularization weight $\lambda \geq 0$ to induce prior knowledge on the desired solution. In this paper, we use the weighted bilateral total variation [21] for sparse and detail-preserving regularization.

## IV. EXPERIMENTS AND RESULTS

This section reports experimental results on five solar module datasets. An example image for every sequence in shown in Fig. 4. The first sequence (D1) exhibits strong noise artifacts while the second sequence (D2) is characterized by heavy lens distortion and streak artifacts. The third sequence (D3) shows substantial motion blur. The last two sequences are not corrupted by any particular kind of artifacts. The sequences have been recorded with four different CMOS cameras at a spatial resolution of $640\,\text{px} \times 512\,\text{px}$. Here, D1, D2 and D3 use different cameras, while D4 and D5 share the same camera.

Furthermore, we have HR images available for the solar module D2-5, which will be used as reference for the crack detection in Sec. IV-D. For D1, there is no such data available, since these images have been captured by a drone on-site.

### IV-A. Reference methods and parametrization

We evaluated multiple SFSR and MFSR methods. For SFSR, we use bicubic upsampling as a baseline and also compare against SRCNN [7] and the recent ESRGAN [11]. For MFSR, we exploit $N = 20$ LR frames and compare our proposed approach against EDVR [27]. Due to the lack of comprehensive solar cell datasets to train deep learning methods for our application, we use the pretrained models provided by the authors for SRCNN, ESRGAN, and EDVR. In contrast, our approach is completely unsupervised and does not require any training.

A qualitative comparison of the results is given in Fig. 5. The DL-based SFSR methods (for example ESRGAN) tend to only sharpen the result and may generate high frequency artifacts as well. Overall, they fail to reconstruct additional details. In contrast, DL-based MFSR (EDVR) only performs well on D3 and D4. Further investigation shows that EDVR

| | segmentation | | | | detection | | |
|---|---|---|---|---|---|---|---|
| Method | precision 3× / 4× | recall 3× / 4× | AP 3× / 4× | $F_1$ 3× / 4× | precision 3× / 4× | recall 3× / 4× | $F_1$ 3× / 4× |
| BICUBIC | 0.57 / 0.55 | 0.09 / 0.09 | 0.22 / 0.22 | 0.15 / 0.15 | 0.59 / 0.58 | 0.21 / 0.22 | 0.31 / 0.32 |
| SRCNN [7] | 0.48 / 0.47 | 0.16 / 0.19 | 0.26 / 0.28 | 0.24 / 0.26 | 0.48 / 0.46 | 0.29 / 0.31 | 0.36 / 0.37 |
| ESRGAN [11] | 0.41 / 0.38 | 0.22 / 0.24 | 0.27 / 0.28 | 0.28 / 0.29 | 0.45 / 0.41 | 0.35 / 0.37 | 0.39 / 0.39 |
| EDVR [27] | 0.37 / 0.33 | 0.31 / 0.36 | 0.33 / 0.33 | 0.34 / 0.34 | 0.40 / 0.34 | 0.42 / 0.45 | 0.41 / 0.39 |
| OURS-H | 0.58 / 0.58 | 0.14 / 0.12 | 0.25 / 0.23 | 0.23 / 0.20 | **0.65** / **0.65** | 0.28 / 0.25 | 0.39 / 0.36 |
| OURS-P | **0.63** / **0.62** | **0.41** / **0.42** | **0.66** / **0.61** | **0.50** / **0.50** | **0.66** / 0.61 | **0.47** / **0.47** | **0.55** / **0.53** |
| Spearman's $\rho$ | 0.49 / 0.66 | 0.60 / 0.31 | 0.60 / 0.31 | 0.60 / 0.31 | 0.49 / 0.60 | 0.60 / 0.31 | 0.77 / 0.14 |
| $p$-value | 0.33 / 0.16 | 0.21 / 0.54 | 0.21 / 0.54 | 0.21 / 0.54 | 0.33 / 0.21 | 0.21 / 0.54 | 0.07 / 0.79 |

**Table I**: Results for the automated segmentation and detection of cracks using the super-resolved cell images. We compare 3× and 4× magnification. The metrics are averaged over D2-5. For the segmentation task, metrics have been computed pixel-wise on the coarse segmentation masks. Here, AP refers to the average precision computed by varying the threshold that is used to compute binary segmentation maps from probability maps. For the detection task, metrics refer to the detection of crack instances, as detailed in Sec. IV. Spearman's $\rho \in [-1, 1]$ is the correlation of the ranking of methods computed from the corresponding metric to the ranking from the user study. Here, $\rho = 1$ indicates perfect agreement, $\rho = 0$ means that the rankings are independent and $\rho = -1$ is achieved, if one ranking is the inverse of the other. The corresponding $p$-values give the possibility that uncorrelated raters result in at least the same amount of inter-observer agreement. A similar test is known from the analysis of statistical significance. Here, small values of $p$ are better, since they indicate that the null hypothesis (raters are uncorrelated) can be rejected.

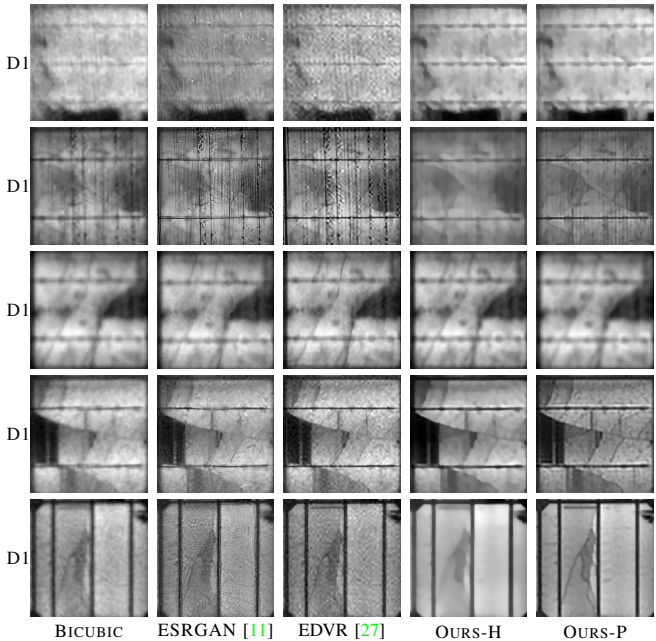

BICUBIC ESRGAN [11] EDVR [27] OURS-H OURS-P

**Fig. 5**: Reconstruction results on D1-5. All images use 4× magnification. We only include ESRGAN for the single-frame super-resolution-methods, since SRCNN produces very similar results.

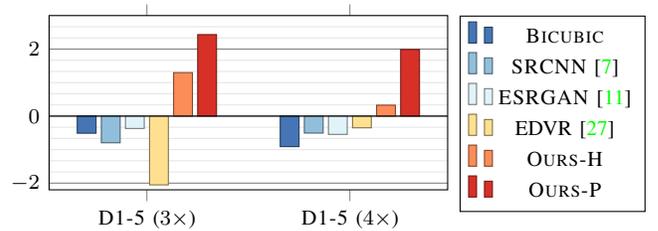

**Fig. 6**: BT-scores from the user study, averaged over D1-5 using 3× magnification (left) and 4× magnification (right). Note that higher scores express better performance.

breaks down completely with stronger noise or artifacts, which is in agreement with the finding of another study in a different context [40]. A detailed comparison of EDVR and our method is given in Fig. 7. In contrast, the proposed method (OURS-P) performs well on most datasets, giving smooth results and showing many details without enhancing noise or artifacts. On D1, OURS-P does not perform well. We found that this is caused by strong motion blur in the LR images that lead to a larger error in the initial parameter estimation using [37] and [38]. In addition, shift-variant motion blur is currently not covered by the reconstruction approach.

### IV-B. Ablation study

For an ablation study concerning the impact of motion estimation and rectification, we replace our registration scheme by standard in-plane homography estimation between planar solar cells using the image alignment toolbox [41]. Then, we determine motion vectors for regularized reconstructions from the homographies. In the following, we denote this method as OURS-H. The impact of our proposed motion estimation can be qualitatively seen from Fig. 5. It turns out that the homography-based approach (OURS-H) gives a relatively blurry result, compared to OURS-P. This is mostly caused by unstable registration.

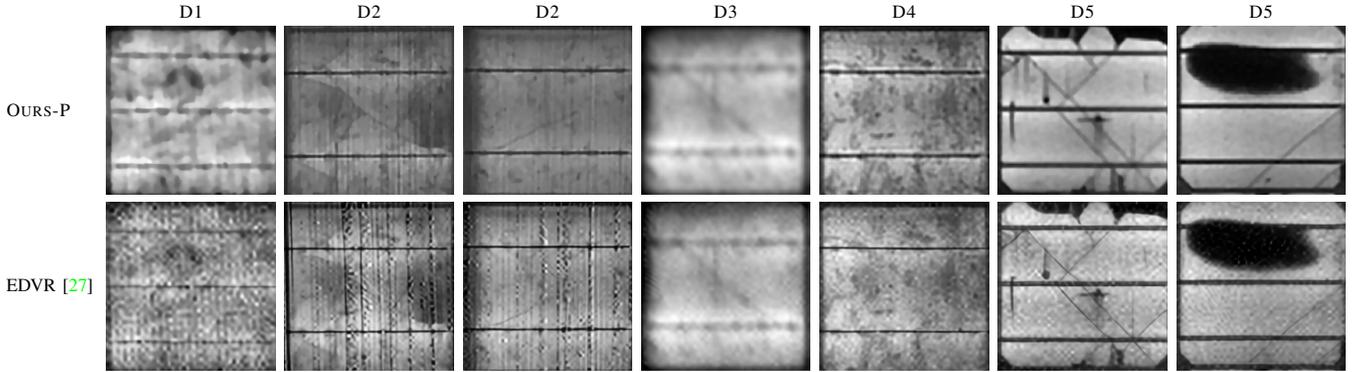

**Fig. 7**: Detailed comparison of the latest DL-based method (EDVR) compared to our method (OURS-P). Here, we report two examples for D2 and D5 and only one for D1, D3 and D4, since the difference is most prominent for D2 and D5.

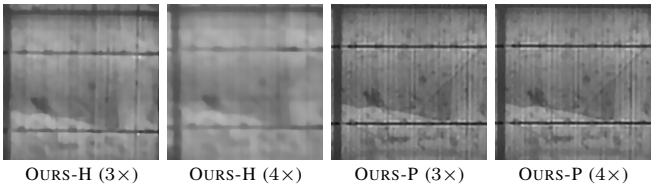

**Fig. 8**: Comparison of $3\times$ vs. $4\times$ magnification for OURS-H and OURS-P.

### IV-C. User study

As opposed to many traditional SR benchmarks, where LR images are obtained by downsampling, we are working with low resolution cameras, such that there are no exact ground truth HR images available. On the upside, this corresponds to a realistic application case. On the downside, we cannot report traditional metrics like PSNR or SSIM that require a pixel-perfect HR reference image. To this end, we conduct a user study. Here, we assess, if SR is suitable for manual inspection of defects by human experts. For the user study, we include all reference methods and our registration in combination with the regularized reconstruction. Further, we include 20 randomly selected cells from every sequence and compare all 6 methods using $3\times$ as well as $4\times$ magnification.

In summary, the user study comprises 3000 pairwise comparisons. It was performed by seven experts in electro-luminescence imaging. They were presented a pair through a web interface and were asked to decide, which of the samples is better suited for detecting defects. During the study, the arrangement of methods (left or right) and the order of samples was randomized. Similar to the benchmark in [13], we use Bradley-Terry (BT) scores [42] derived from pair-wise comparisons to compute a ranking of the methods. Using the scores $p_k$ and $p_l$, the probability that method $k$ is preferred over method $l$ is computed as $P(k > l) = \exp(p_k)/(\exp(p_k) + \exp(p_l))$.

The result is presented in Fig. 6. On average, the DL approaches do not perform well. In contrast, OURS-H shows moderate performance, but is limited by instable registration. Our proposed method clearly outperforms the other methods. Using the BT scores, we find that OURS-P is preferred over OURS-H with a probability of $P = 0.57$ ($3\times$) or $P = 0.61$ ($4\times$). Regarding the magnification factor, it turns out that the difference in perceived performance is increased for $4\times$ magnification. This is explained by the negative impact of instable registration for the OURS-H approach, which is amplified with the increased magnification (see Fig. 8).

Finally, we evaluate the inter-observer variance to determine the amount of agreement between users. We compute the ranking of methods per user and Kendall's coefficient of concordance $W \in [-1, 1]$ to obtain the inter-observer variance of method rankings, where $W = 1$ denotes total agreement between users, $W = 0$ denotes no agreement and $W = -1$ states that user rankings are the exact inverse of each other. We find that the value of $W$ depends on the magnification. For $3\times$ magnification, the coefficient computes as $W_{3\times} = 0.33$, whereas for $4\times$ magnification the agreement is stronger and amounts to $W_{4\times} = 0.44$. The same effect has been reported by Köhler *et al.* in a similar study [13]. We conclude that, with higher magnification factors, different methods are easier to distinguish for human observers, which leads to a higher overall consensus.

### IV-D. Automated crack segmentation and detection

We also investigate if the proposed method improves automated inspection. For this analysis, we include all cell images from D2-5, since HR measurements are available for this data. The results are summarized by Tab. I and Fig. 9.

For crack segmentation, we apply the weakly supervised approach proposed by Mayr *et al.* [43] to the results with $3\times$ and $4\times$ magnification as well as to the HR images. We use the segmentation result obtained on the HR images as ground truth. Obviously, this automatically generated ground truth is not perfect. However, the bias is constant for every method. Hence, the relative performance of the methods remains unaffected.

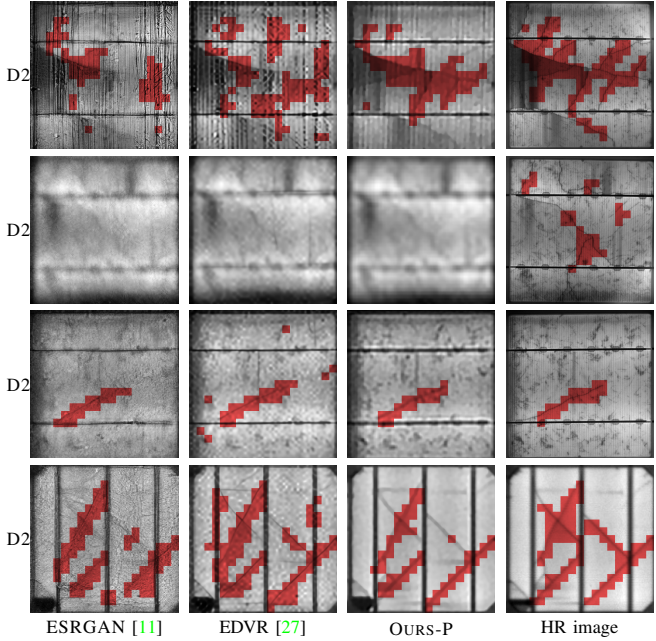

**Fig. 9**: Coarse automatic crack segmentation results of the three methods with the highest segmentation $F_1$ score using images from D2-5. All images use $4\times$ magnification.

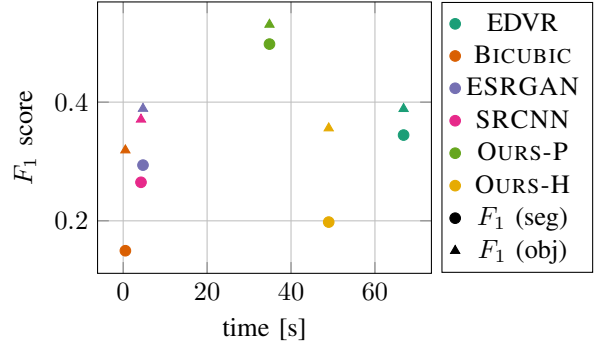

**Fig. 10**: Comparison of the methods regarding computation time and segmentation metrics. Except for the bicubic baseline, all results have been computed on GPU. Most methods involve additional preprocessing, which is not included with the results, since it is not consistently implemented on GPU and the computation time is dominated by the reconstruction step anyway.

Mayr *et al.* use a fully convolutional network trained for classification of cracks. The segmentation is then derived from the class activation maps. As a result, the resolution of segmentation maps is reduced by a factor of 16, compared to the original image. This is beneficial to our task, since a registration of HR images that are used as pseudo ground-truth with the super-resolved cell images is challenging and the registration mismatch is hidden by the subsampling.

We report the pixel-wise metrics with respect to the automated segmentation obtained from the HR images. It turns out that EDVR results in a recall that is comparable to our method. However, it generates many false positives, as indicated by the low precision and average precision (AP) in Tab. I. On the other hand, OURS-H results in a high precision but a small recall. Overall, our method performs $\sim 2\times$ better than EDVR and $\sim 3\times$ better than BICUBIC regarding the average precision and $F_1$ score.

In addition to the pixel-wise metrics, we compare the detection of crack instances too. For many applications, it is more interesting to know the number of cracks, rather than having a perfect segmentation. Furthermore, a segmentation that is only one pixel off results in a significant drop in the quantitative segmentation performance, since cracks are usually thin.

We perform connected component analysis on the coarse segmentation masks, assuming that a single connected component corresponds to a crack instance. Then, we compute the fraction of crack instances for the HR segmentation masks, where at least one pixel of the corresponding area on the SR image is classified as a crack. This is the recall of the crack instance detection. Inversely, the fraction of detected crack instances on the SR image that overlaps with a crack instance detected on the HR image by at least a single pixel, is the precision of crack instance detection. We find that our method outperforms the other methods with respect to recall and $F_1$ score. OURS-H performs a little better regarding the precision, but fails in the recall of crack instances. Furthermore, EDVR has a recall that is close to ours, but has a much lower precision. Regarding the $F_1$ score, which combines recall and precision into a single score, our method performs $\sim 35\%$ better than the best reference method.

Finally, we analyze the correlation between the results obtained from the user study to the quantitative results. Spearman's rank correlation coefficient $\rho \in [-1, 1]$ is widely used to quantify the amount of correlation between rankings of ordinal variables. As opposed to, for example, Pearson's correlation, it is not limited to linear associations. The results are summarized by Tab. I. We find that, apart from the precision score, the association between the metrics and the user study rankings is weaker for the higher magnification. This complies to a previous study, that reported a weaker correlation for reference metrics at higher magnifications, as opposed to reference-less metrics [13]. Overall, the experiment reveals a fair correlation of the user study results to the quantitative results for $3\times$ magnification, while they do not correlate well for $4\times$ magnification. This shows that, for higher magnifications, qualitative analysis of the results cannot substitute the qualitative analysis and vice versa.

## IV-E. Computational performance

From an application perspective, not only quality of results, but also runtime matters. To this end, we conduct an evaluation of the computational performance. The experiments have been conducted using a workstation equipped with an Intel i7-7820X CPU, $64\,\mathrm{GB}$ or RAM and a Nvidia Titan Xp GPU using a magnification of $4\times$ with an initial resolution of $512^2$ pixels. Note that the results were averaged over D1-5. In Fig. 10, we summarize the computational performance and compare it to the quantitative segmentation metrics. In summary, we find that, in our experiments, SFSR methods are much faster than MFSR and might be suitable for online image processing. However, we have shown that they are not well suited for the task at hand, since HR details are not authentic. In contrast, MFSR methods require about $30\,\mathrm{s}$ to $50\,\mathrm{s}$ of processing time, with the proposed method being faster than EDVR. As a result, MFSR needs to run offline. This is not prohibitive for many applications: For example, given a scenario, where data is collected on-site and later analyzed to identify and possibly replace faulty modules, this is an acceptable amount of computation time.

## V. CONCLUSION

This paper presents a new method to increase the spatial image resolution of solar modules. It is based on parametric registration and can be combined with various reconstruction approaches. As opposed to non-parametric registration, such as optical flow, this ensures that the registration result is plausible given the imaging process. In contrast to parametric registration based on plain homography estimation, it accounts for lens distortion, too.

We present a user study showing that our approach outperforms state of the art single-frame and multi-frame super-resolution methods. Furthermore, we combine the proposed method with automatic crack detection and segmentation and show that SR is not only beneficial for manual inspection by human experts, but also for automated inspection. In particular, we show that the proposed method is better suited for subsequent automated inspection than state of the art methods. Furthermore, we investigate the methods in terms of computational performance and find that the proposed method is computationally fast, compared to other state of the art MFSR methods.

In future works, we aim to integrate the proposed motion model with SR to form a single optimization problem, as shown for the purely projective case in [33]. This would also allow for joint estimation of motion. Furthermore, an interesting opportunity is to combine the conventional and DL-based methods in terms of known operator learning [44] to join the robustness of conventional methods with the power of learning individual steps of the algorithm from the data.

**Acknowledgements** We gratefully acknowledge funding of the Federal Ministry for Economic Affairs and Energy (BMWi: Grant No. 0324286, iPV4.0). In addition, we would like to thank the user study participants: J. Hepp, M. Mayr, A. Vetter, J. Denz, T. Winkler and C. Buerhop. Further, we would like to thank Ircam GmbH and Greateyes GmbH for providing the cameras and images and Rauschert Heinersdorf-Pressig GmbH for access to their photovoltaic plants.

# APPENDIX A
## DERIVATION OF INVERSE DISTORTION

Let $r_d$ denote the distorted and $r_u$ denote the undistorted radius. Then, the distorted radius using first-order radial

symmetric distortion is given by

$$r_d = r_u + \kappa r_u^3 . \quad (8)$$

Substitute $z = 3\kappa r_u$ and set $p = 3\kappa$ and $q = -27\kappa^2 r_d$ to obtain

$$z^3 + 3pz + q = 0 . \quad (9)$$

The discriminant is given by $D = q^2 + 4p^3$. Depending on the discriminant, there are different solutions:

For $D > 0$, there is one real solution:

$$z_1 = \frac{1}{2}\sqrt[3]{-4q + 4\sqrt{q^2 + 4p^3}} +$$
$$\frac{1}{2}\sqrt[3]{-4q - 4\sqrt{q^2 + 4p^3}} . \quad (10)$$

For $D = 0$, it holds that $27^2 \kappa^2 r_u^2 = 108\kappa^3$, which has one real root at $\kappa = 0$. Therefore, if $D = 0$, the solution is given by

$$r_u = r_d . \quad (11)$$

For $D < 0$, there are three different real solutions:

$$z_n = 2\sqrt{-p}$$
$$\cos\left(\frac{1}{3}\arccos\left(\frac{-q}{2\sqrt{-p^3}}\right) + \frac{(n-1)\cdot 2\pi}{3}\right)$$
$$1 \leq n \leq 3 . \quad (12)$$

We require that $r_d$ is continuous at $\kappa = 0$ for fixed $r_u$ and find by experiment that this is only true for $z_3$.